\documentclass[10pt,twocolumn,letterpaper]{article}

\usepackage{cvpr}
\usepackage{times}
\usepackage{epsfig}
\usepackage{graphicx}
\usepackage{amsmath}
\usepackage{amssymb}
\usepackage{subcaption}
\usepackage{mwe}
\usepackage{float}
\usepackage{dirtytalk}
\usepackage{enumitem}
\usepackage{adjustbox}
\usepackage[toc,page]{appendix}

\graphicspath{{/Images}}

\usepackage[bookmarks=false]{hyperref}
\hypersetup{breaklinks=true}

\cvprfinalcopy 

\def\cvprPaperID{****} 

\begin{document}
\title{Detection and Segmentation of Custom Objects using High Distraction Photorealistic Synthetic Data}
\author{\bf Roey Ron\\
DataGen\\
{\tt\small roey.ron@datagen.tech}
\and
\bf Gil Elbaz\\
DataGen\\
{\tt\small gil@datagen.tech}
}

\maketitle

\begin{abstract}
   We show a straightforward and useful methodology for performing instance segmentation using synthetic data. We apply this methodology on a basic case and derived insights through quantitative analysis. We created a new public dataset: The $Expo\ Markers\ Dataset$ intended for detection and segmentation tasks. This dataset contains 5,000 synthetic photorealistic images with their corresponding pixel-perfect segmentation ground truth. 
   The goal is to achieve high performance on manually-gathered and annotated real-world data of custom objects. We do that by creating 3D models of the target objects and other possible distraction objects and place them within a simulated environment.
   
   Expo Markers were chosen for this task, fitting our requirements of a \textit{custom object} due to the exact texture, size and 3D shape. An additional advantage is the availability of this object in offices around the world for easy testing and validation of our results. We generate the data using a domain randomization technique that also simulates other photorealistic objects in the scene, known as \textit{distraction objects}. These objects provide visual complexity, occlusions, and lighting challenges to help our model gain robustness in training. We are also releasing our manually-gathered datasets used for comparison and evaluation of our synthetic dataset.\newline
   This white-paper provides strong evidence that photorealistic simulated data can be used in practical real world applications as a more scalable and flexible solution than manually-captured data. Code is available at the following address:
   \url{https://github.com/DataGenResearchTeam/expo_markers}
\end{abstract}

\begin{figure}[ht!]
\centering
\renewcommand{\arraystretch}{0}
\setlength{\tabcolsep}{0pt}
\begin{tabular}{c c}
\includegraphics[width=0.5\linewidth]{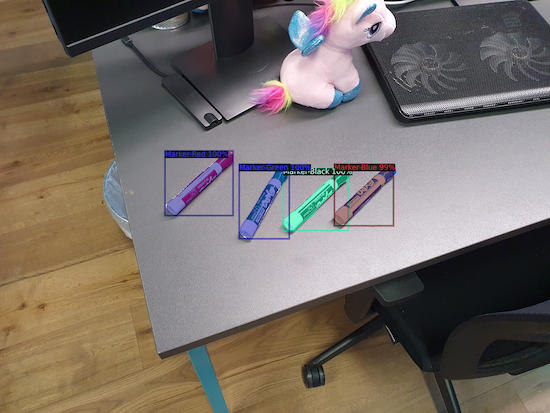} & 
\includegraphics[width=0.5\linewidth]{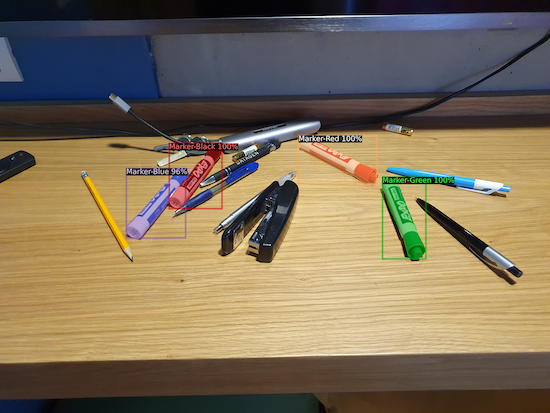} \\
\includegraphics[width=0.5\linewidth]{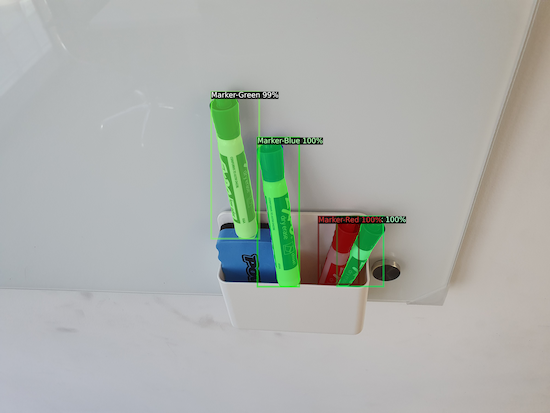} & 
\includegraphics[width=0.5\linewidth]{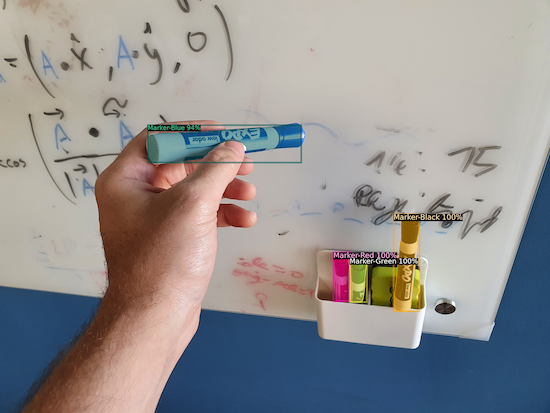}
\end{tabular}
    \caption{Prediction samples on $Real\ A$. The model was trained purely from synthetic data.}
    \label{fig:markers_synt}
\end{figure}


\section{Introduction}
Nowadays, supervised deep learning algorithms, more specifically Convolutional Neural Networks, outperform classical machine learning and computer vision algorithms on all standard tasks. These algorithms reach state-of-the-art performance, but at the same time require large amounts of labeled data to provide a robust solution for real-world scenarios~\cite{o2019deep}. Data collection and labeling is an expensive, time-consuming process, and the quality of the data varies between providers, especially in pixel-wise tasks such as segmentation. In some tasks, it is almost impossible for humans to extract the labels (such as depth maps or surface normal maps). Sun \etal \cite{sun2017revisiting} discuss the success of deep learning as a function of dataset size and predict that lack of labeled data will remain a bottleneck for improving results. Sun point out that, while GPU computation power and model sizes have continued to increase over the last years, the size of the largest training datasets has surprisingly remained constant. They also show that the performance on vision tasks increases logarithmically with the volume of training data, and that higher capacity models are better at efficiently using large datasets. 

For many computer vision applications, including autonomous driving, security, smart store and interactive robotics, the end-user expects high-quality results that are robust and reliable. For these applications, using the most proven and highest quality supervised machine learning algorithms after training them with large amounts of data is the go-to solution. While few-shot learning, weakly supervised machine learning, unsupervised machine learning or deep feature extraction, are popular topics and are showing promising results, supervised learning is usually simple, effective and easy to train.


\subsection{Synthetic Datasets}
The promise of synthetic data has been clear since its inception; datasets generated through computational algorithms that can mimic semantic and visual patterns found in the real world. This data could train machine learning algorithms without compromising privacy (e.g. facial data) or being susceptible to high-level biases in the data (e.g. generation of equal amounts of males and females in the dataset). Synthetic data is also highly scalable -- more data could always be generated -- and edgecases -- gaps in the data where manually-gathered data would be hard or unreasonable to collect -- could be generated.
In the past years, synthetic data has shown promise across a range of verticals, from medical research, where patient privacy is high priority, to fraud detection, where synthetic datasets can be used to test and increase the robustness of security solutions.
In recent years, synthetic data generation has gained substantial popularity within the computer vision field~\cite{ros2016synthia, jalal2019sidod, kar2019meta} as a solution to the data bottleneck problem. Two main approaches bridge the gap between the source domain (synthetic data) and the target domain (manually-gathered) -- domain randomization and domain adaptation.

\subsection{Domain Randomization}
The core idea behind Domain Randomization (DR) is to train a neural network to perform well on such a broad synthetic source domain that the model will generalize to the real-world target domain data. This method is relevant when the exact target domain is unknown, highly variant or hard to mimic. The scene is randomized in non-realistic ways to force the neural network to learn in two ways. First, it will learn the robust features of the scenes that are invariant in all randomizations. For example, if we randomize the lighting in the scene but keep the geometric structures constant, the network will become more robust to unforeseen lighting situations. Second, the random sub-domains generated by the DR that are closer to the target domain will teach the neural network to analyze the target domain, without explicitly recreating it. For instance, ~\cite{tremblay2018training, tobin2017domain} apply DR techniques both to the scene structure (placing the objects) and to the textures. They use general basic objects (e.g. cube or pyramid without predefined texture) to create the variance they needed. In our case, we use photorealistic objects.

\subsection{Domain Adaptation}
Several works ~\cite{Zhu_2017, liu2019few, hoffman2017cycada, li2018semantic} deal with the task of transferring images from a source visual domain to another separate target visual domain in order to attempt to mimic the effectiveness of standard, collected and annotated visual data, captured from the target domain. By closing the domain gap, the synthetic data can theoretically act as if it was captured from the target domain.
Domain Adaptation can also be applied by adjusting the data to a specific camera hardware, by using small amount of unlabeled images captured with the target camera. In our work, we use photorealistic images (in terms of image style, not content). By doing so, we minimize the visual domain gap. In future works we will explore the use of domain adaptation to further improve our results.

\section{Contributions}

\subsection{Datasets}
\begin{figure*}[!ht]
    \centering
    \setkeys{Gin}{height=1.3in}
\begin{subfigure}{0.4\linewidth}
    \centering
    \caption*{$Synthetic$}
\includegraphics{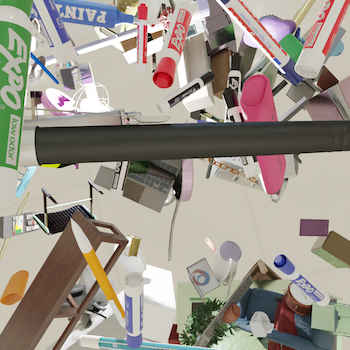} \includegraphics{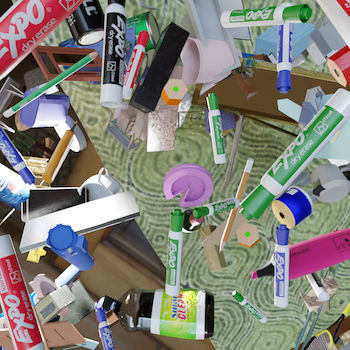}\\
\includegraphics{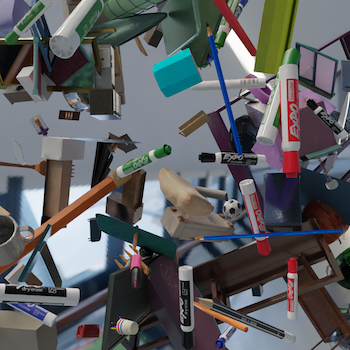} \includegraphics{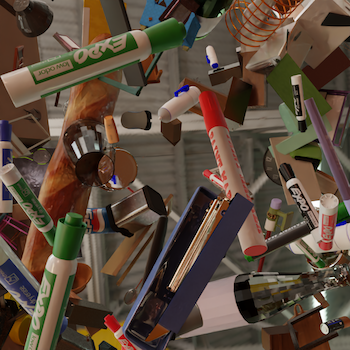}
\end{subfigure}
\hfil
\begin{subfigure}{0.29\linewidth}
    \centering
    \caption*{$Real\ A$}
\includegraphics{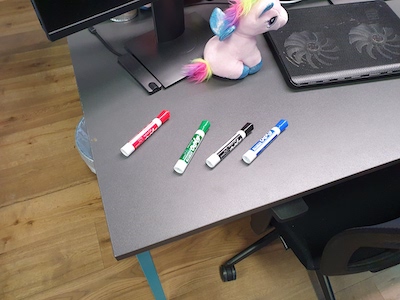}\\
\includegraphics{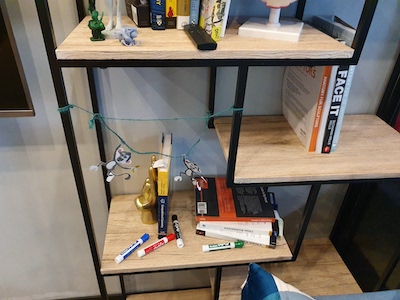}
\end{subfigure}
\hfil
\begin{subfigure}{0.29\linewidth}
    \centering
    \caption*{$Real\ B$}
\includegraphics{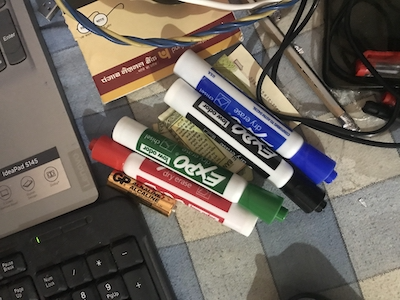}\\[0pt]
\includegraphics{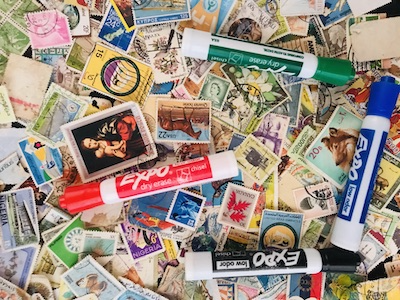}
\end{subfigure}
\caption{$Expo\ Markers\ Dataset$.}
\label{fig:expo_datasets}
\end{figure*}

In this paper we used three datasets (Table \ref{table:expo_datasets}), Figure \ref{fig:expo_datasets} shows sample from the three datasets next to each other.

\begin{table}[H]
\centering
\resizebox{\columnwidth}{!}{%
\begin{tabular}{ c | c | c | c } 
Dataset Name & Size & Type & Usage \\ 
\hline
$Synthetic$ & 5,000 & Synthetic & Training\\
$Real\ A$ & 250 & Manually-Gathered & Evaluation\\
$Real\ B$ & 1,000 & Manually-Gathered & Training and Evaluation\\
\end{tabular}
}
\caption{$Expo\ Markers\ Dataset$. During our experiments, we used the datasets in three ways: training, validation and test. Each data point was used for one purpose only.}
\label{table:expo_datasets}
\end{table}%

\subsubsection{Synthetic Dataset}

The synthetic dataset - $Synthetic$ consists of 5,000 synthetic photorealistic images with their corresponding pixel-perfect segmentation ground truth. Image resolution is 1024x1024 and each image contains, on average, 13 marker object instances and 80 distractor object instances. Figure \ref{fig:synt_samples} presents samples from the $Synthetic$ dataset.

\subsubsection{Manually-Gathered Datasets}
In order to compare and evaluate the performance of our synthetic data, we have collected two manually-gathered datasets: $Real\ A$ and $Real\ B$, these two datasets were taken with different devices and in different environments, and we consider them as two different domains. In both of the datasets, some of the images are quite simple, while others are more complicated and include occlusions and objects which are similar to the markers (e.g., pens, pencils, and other brands of markers). The vast majority of the manually-gathered images contains 4 markers, one in each marker color.
$Real\ A$ was captured with Samsung s10+ camera, the dataset consists 250 images and is used for evaluation.
$Real\ B$ was captured with Apple iPhone 7 camera, the dataset consists 1,000 images and is used for both training and evaluation.
Samples from the datasets $Real\ A$ and $Real\ B$ are presented in Figures \ref{fig:office_samples} and \ref{fig:india_samples} respectively.

\subsection{Code}
We share our code which enables 'plug and play' inference and training, using Mask R-CNN \cite{he2017mask} on our dataset. Our code is based on detectron2 \cite{wu2019detectron2}, which is fast, flexible and enables the use of various architectures (see detectron2's `model zoo': \href{https://github.com/facebookresearch/detectron2/blob/master/MODEL_ZOO.md#coco-instance-segmentation-baselines-with-mask-r-cnn}{\color{blue}{detectron2/MODEL\_ZOO.md}}).

\section{Method}

\subsection{Data Generation}

First, we created a photorealistic 3D representation of our target class - the Expo markers. These representations were created by 3D artists using accurate modeling.

In addition, a set of photorealistic items were used as distraction for the algorithm. These items were queried at random from the indoor environment section of the DataGen Asset Library, a library of hundreds of thousands of photorealistic items. 

The targets were placed in the 3D scene, visible to the simulated camera lens. The distraction items were placed within the same 3D scene in order to create a challenging visual setting. The number of objects, backgrounds, scene lighting, occlusions and randomness of orientation do not attempt to resemble the distributions in a real-world scene, and instead try to provide the algorithm a more challenging dataset to train on. The main goal is to enable the algorithm to learn a robust representation that could face extreme cases present in the real world.

Each image was rendered with Cycles rendering engine \cite{cycles} and a segmentation map was created. This method outputs photorealistic images with pixel level annotations.

Using the DataGen dataset generation tool, we created a synthetic dataset of 5000 images. A few calibration iterations were required to set reasonable parameter values for the distributions over the objects placement, lighting conditions, and number of appearances.

The ease of generating synthetic data (after the initial effort of creating the data generation tool) enabled us to quickly perform multiple iterations using different parameters until we achieved satisfactory results.

\subsection{Training and Evaluation}
Mask R-CNN~\cite{he2017mask} was used for training and evaluation. for each training session, we used COCO's initial weights, mini batch SGD as optimization algorithm with mini batch size of 4. Detectron2's default augmentations were used, the network was trained for 40k iterations and a weights file was saved every 500 iterations.
The initial learning rate of 0.003 was reduced twice: at 30k and 37k iterations, by a factor of 0.1. Since our initial weights (COCO) were trained on manually-gathered data, and our training data is partly synthetic or purely synthetic in some  of the setups, there is more interest in monitoring if and where over-fitting to synthetic data arises during the training process. Best weights were chosen by evaluating the network's $mAP$ on a validation set of size 50 (from the same visual distribution as the test set) on each of the weights that were saved during the training session after every 500 iterations. Later, the result metric values achieved by evaluating the best chosen weights on a test set of size 200. Each experiment was performed 5 times and the average results were reported.

\subsection{Performance Metrics}
mAP (Mean Average Precision) is used as our main performance metric.
For convenience, we define some notation that we will use to discuss our results:
\setlist{nolistsep}
\begin{itemize}[noitemsep]
\small
\itemsep0em 
\item$mAP$ - mAP at Intersection over Union IoU=.50:.05:.95 (COCO primary metric).
\item$mAP^{IoU=.50}$ -  mAP at IoU=.50 (PASCAL VOC metric). 
\item$mAP^{IoU=.75}$ - mAP at IoU=.75 (strict metric).
\item$mAR$ - (Mean Average Recall) for 100 detections per image.
\end{itemize}

\section{Results}

\subsection{Quality Analysis of Synthetic Data} 
\begin{figure}[h!]
    \centering
    \includegraphics[width=8cm]{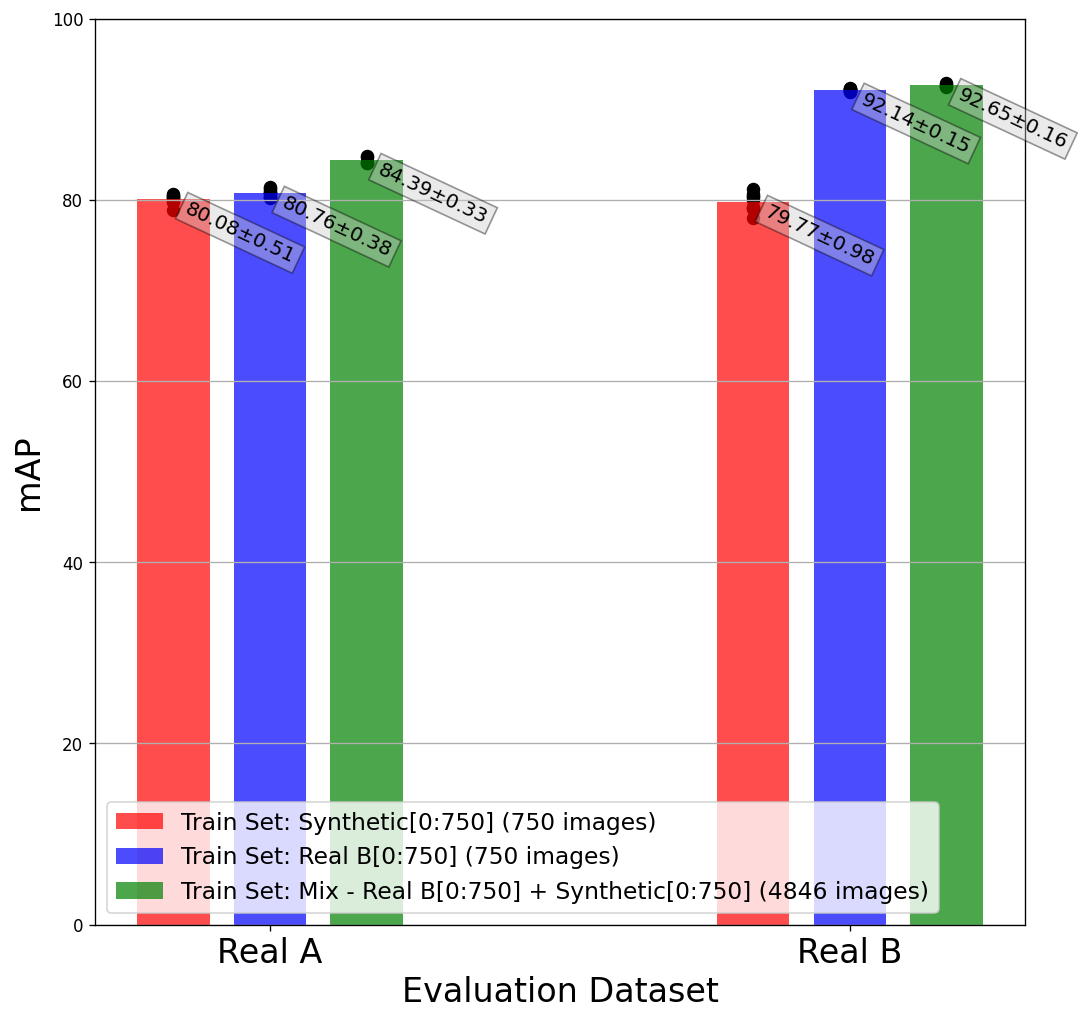}
    \caption{Quality Analysis of Synthetic Data - Mask R-CNN was trained on three datasets: Synthetic, Manually-Gathered and Mix. Each of the networks was evaluated on two datasets: $Real\ A$ and $Real\ B$}.
    \label{fig:main_results}
\end{figure}

A qualitative comparison between three training datasets: purely synthetic, purely manually-gathered and a mix of both is presented in Figure \ref{fig:main_results}. Each training dataset was evaluated on two datasets: $Real\ A$ and $Real\ B$. The two evaluation datasets provide an example of two different real world scenarios. The first evaluation setup on $Real\ A$ can tell us about the use case, in which a small amount of manually-gathered training data is available. But, it comes from a slightly different domain than the target domain, covers a small portion of the target visual domain or is captured with a different device, to the target capture device. In this case we observe that the synthetic data - $Synthetic$ and the manually-gathered data - $Real\ B$ achieve similar performance in terms of $mAP$ value and that their mixture achieves significantly improved results. The second evaluation setup on $Real\ B$ shows us the use case where manually-gathered training data is available from the same domain, captured from the same device and the same scenario as a narrow target domain. We observe that the synthetic data achieves satisfactory results, while the manually-gathered data from exactly the same distribution achieves better results than the synthetic data. However, even in this perfect use case, there is an improvement when training with a mixture of manually-gathered and synthetic data.

\subsection{Performance as a Function of Dataset Size}

\begin{figure*}[ht]

\begin{subfigure}[h]{0.49\linewidth}
\includegraphics[width=\linewidth]{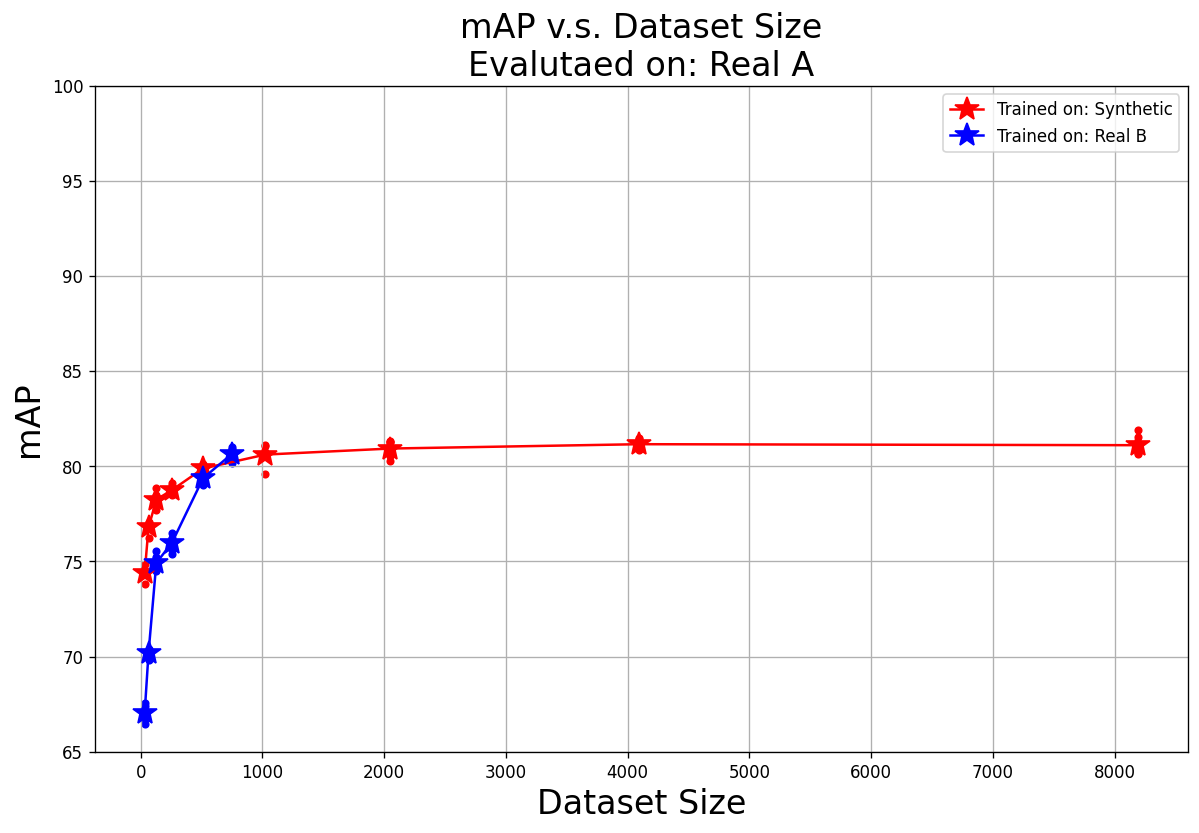}
\caption{}
\end{subfigure}
\hfill
\begin{subfigure}[h]{0.49\linewidth}
\includegraphics[width=\linewidth]{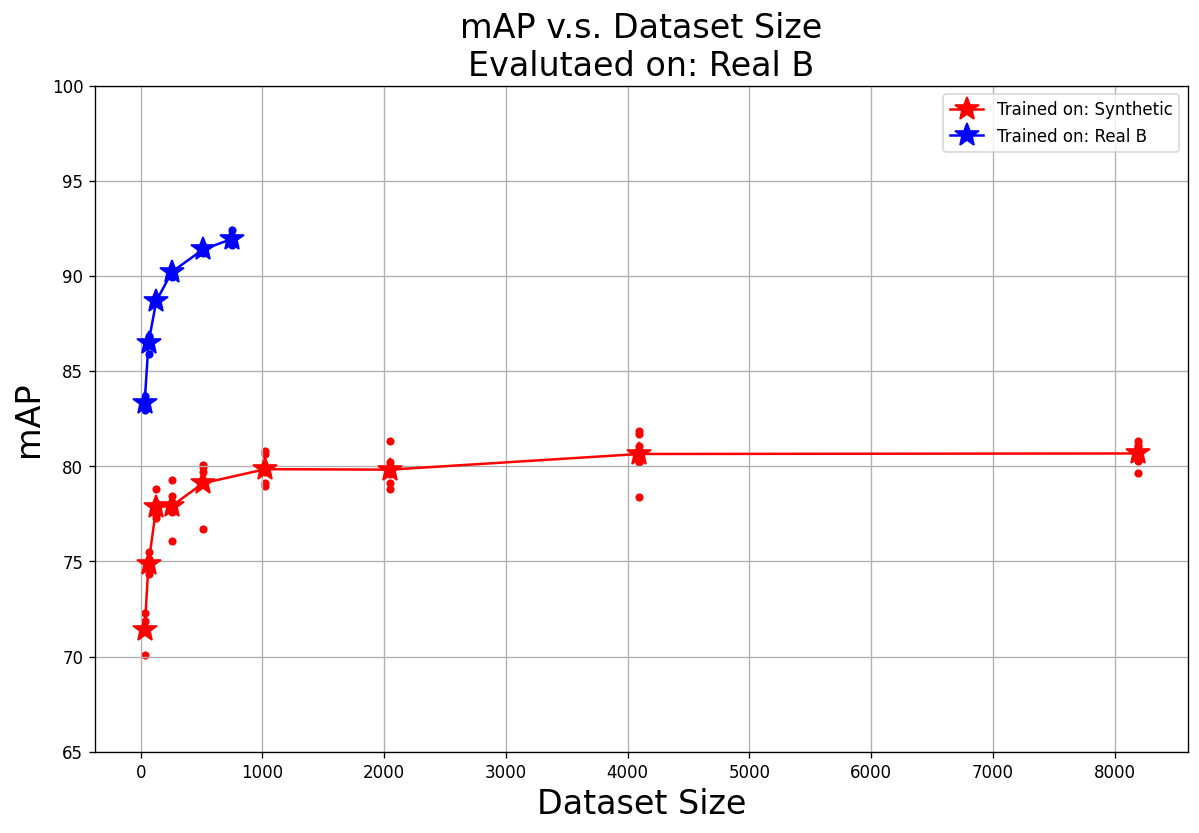}
\caption{}
\end{subfigure}%

\caption{Performance as a Function of Dataset Size: (a) Evaluated on $Real\ A$ (b) Evaluated on $Real\ B$.}
\label{fig:map_vs_size}
\end{figure*}

In order to understand how the dataset size affects the performance when training on synthetic and manually-gathered data, we trained Mask R-CNN with synthetic and manually-gathered data and evaluated the trained networks on both of the manually-gathered datasets. Results can be seen in Figure \ref{fig:map_vs_size}. When training with synthetic data, we used the following dataset sizes: [32, 64, 128, 256, 512, 1024, 2048, 4096], while when training with the limited sizes manually-gathered data we used the following dataset sizes: [32, 64, 128, 512, 750].




\subsection{Training with Mixture of synthetic data and Manually-Gathered}
\begin{figure*}[ht!]
        \centering
        \begin{subfigure}[b]{0.49\textwidth}
            \centering
            \includegraphics[width=\textwidth]{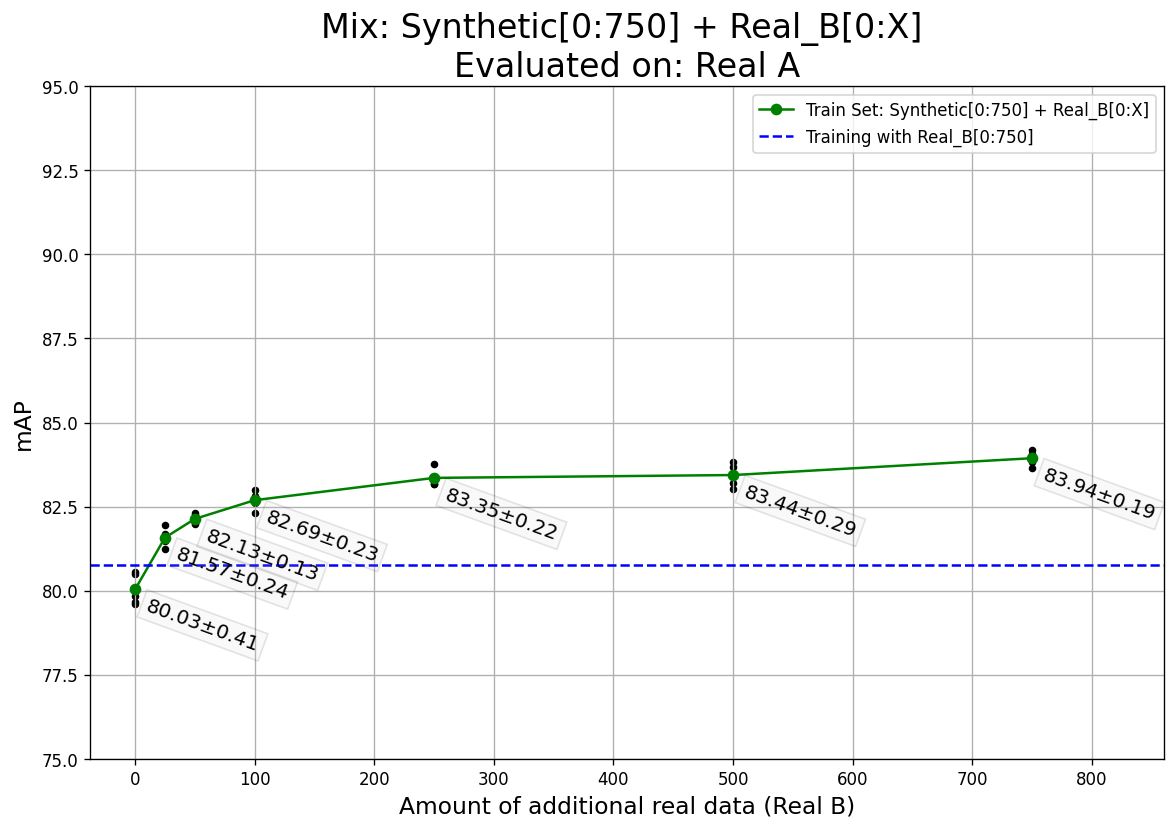}
            \caption[]%
            {{\small }}    
            \label{fig:mean and std of net14}
        \end{subfigure}
        \hfill
        \begin{subfigure}[b]{0.49\textwidth}  
            \centering 
            \includegraphics[width=\textwidth]{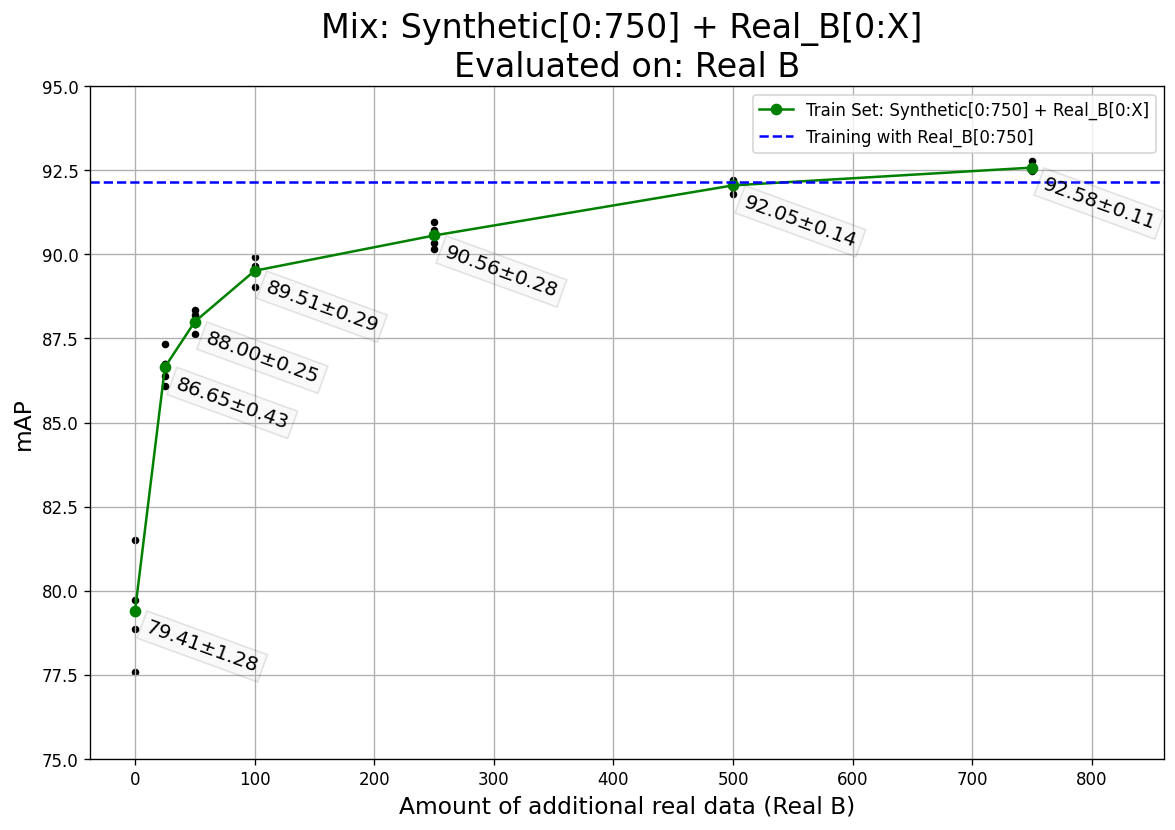}
            \caption[]%
            {{\small }}    
            \label{fig:mean and std of net24}
        \end{subfigure}
        \vskip\baselineskip
        \begin{subfigure}[b]{0.49\textwidth}   
            \centering 
            \includegraphics[width=\textwidth]{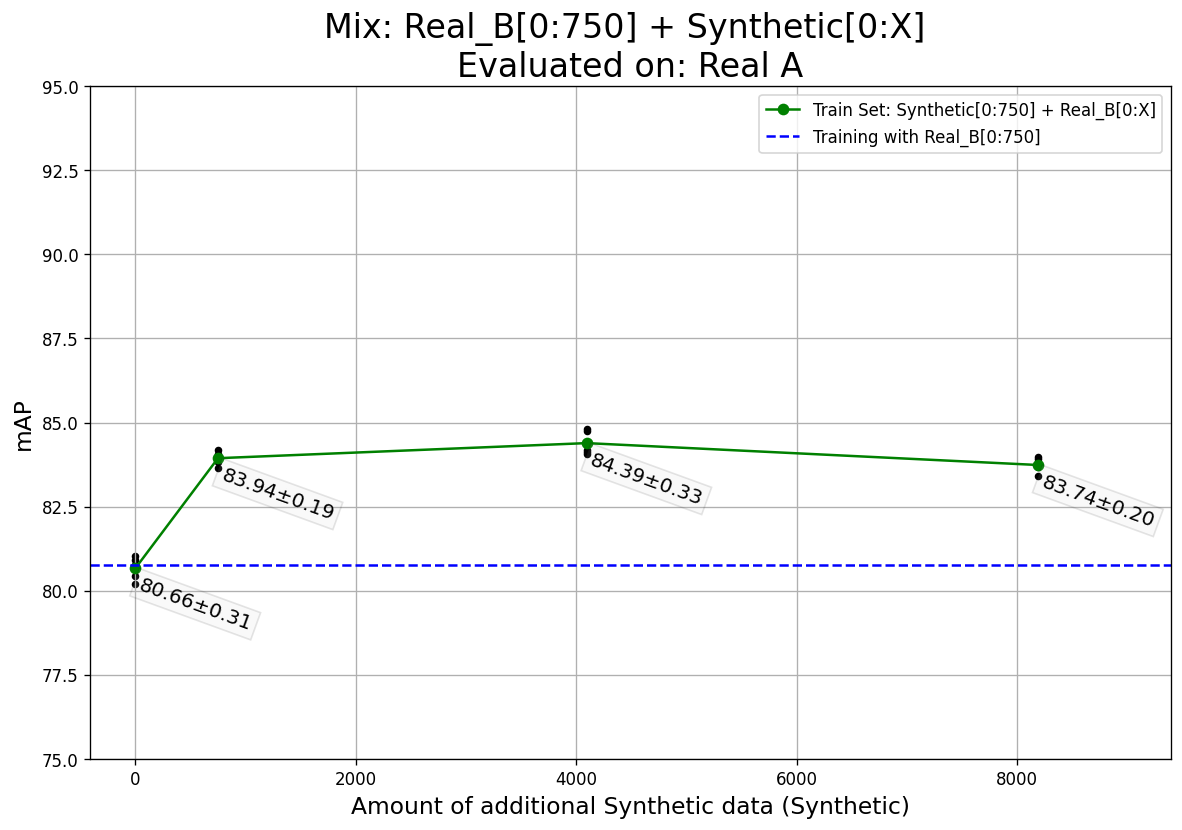}
            \caption[]%
            {{\small }}    
            \label{fig:mean and std of net34}
        \end{subfigure}
        \hfill
        \begin{subfigure}[b]{0.49\textwidth}   
            \centering 
            \includegraphics[width=\textwidth]{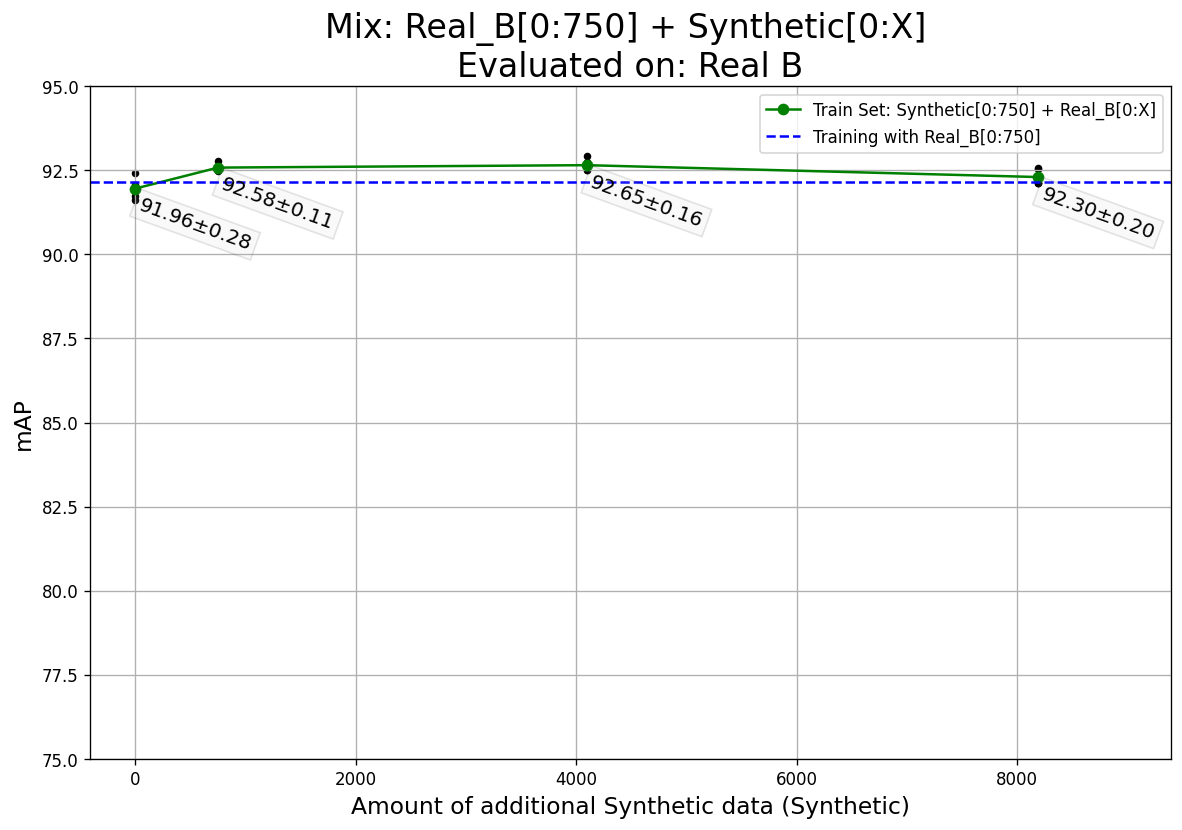}
            \caption[]%
            {{\small}}    
            \label{fig:mean and std of net44}
        \end{subfigure}
        \caption[ The average and standard deviation of critical parameters ]
        {\small Training with different mixtures of synthetic data. (a)+(b): training with fixed amount of synthetic data while changing the amount of manually-gathered data. (c)+(d): training with fixed amount of manually-gathered data while changing the amount of synthetic data.} 
        \label{fig:mix_fig}
    \end{figure*}

Mask R-CNN~\cite{he2017mask} was trained on different mixtures of $Synthetic$ and $Real\ B$ and was evaluated both on $Real\ A$ and $Real\ B$, the results are presented in Figure \ref{fig:mix_fig}.

\subsection{Mask R-CNN on Different Datasets}
In order to have some reference points to our results and to have some context and sense of the Mask R-CNN capabilities we show results of MASK R-CNN with similar backbones trained on different datasets. Mask R-CNN~\cite{he2017mask} was originally trained on a subset of the COCO dataset~\cite{lin2014microsoft} (which contains 80 classes) using the union of 80k training images and a 35k subset of validation images. 

\begin{table}[H]
\centering
    \begin{tabular}{c | c | c | c c c } 
    
     Train Set & Test Set & Backbone & AP & AP$_{50}$ & AP$_{75}$ \\ 
     \hline
     COCO  & COCO & 50-FPN & 33.6 & 55.2 & 35.3 \\ 
     COCO & COCO & 101-FPN & 35.4 & 57.3 & 37.5 \\ 
    Meta-Sim & KITTI & 50-FPN & N/A & 77.5 & N/A \\
    Ours & Ours & 101-FPN & 79.2 & 97.5 & 95.2
    
    \end{tabular}
\caption{Performance of Mask R-CNN on different datasets. 50-FPN and 101-FPN stands for ResNet-50-FPN and ResNet-101-FPN respectively}
\label{table:maskrcnn_paper_results}
\end{table}%


In Table~\ref{table:maskrcnn_paper_results}, we can see that using ResNet-101-FPN trained on COCO, Mask R-CNN achieves $mAP=33.6$ while we managed to achieve $mAP=79$. That said, this comparison is asymmetrical for two main reasons: first, our dataset contains 4 classes while there are 80 classes in COCO. Second, the notion of class is quite different, the classes in COCO are broader. For instance, there is a ‘keyboard’ class in COCO which includes many types of different keyboards, while in our dataset each class contains exactly one specific object.

Kar \etal~\cite{kar2019meta} generated a synthetic dataset and trained Mask R-CNN with Resnet-50-FPN as their backbone. Their target domain was the KITTI dataset and they applied domain adaptation. They achieved $mAP^{IoU=.50}=0.77$ on the KITTI dataset, with the easiest setup.

Comparison of this data and methodology will continue. This initial step allows us to show promising results that are relevant for real-world applications while only training on the synthetic data domain. 


\section{Discussion}

\subsection{Quality Analysis of Synthetic Data}
We see from the qualitative analysis that synthetic data has satisfactory results generalizing to our two real domains.

\subsection{Dataset Size Impact on Performance}
The relation between performance and data set size was tested and is described in Figure~\ref{fig:map_vs_size}. From these results it is observable that the performance increases as a function of dataset size. The result fits well the claims made by ~\cite{sun2017revisiting}, in which, larger datasets result in better performance. 

It is noticeable that even small amounts of data achieve satisfactory results, this fact may tell us that the Expo Markers Domain is essentially a relatively simple toy domain. We also observe in Figure \ref{fig:map_vs_size}(a) that the synthetic data reach comparable or even better results than the manually gathered data. Even though more manually gathered data may perform better, we were bounded by the amount of manually gathered data we had. This is a realistic scenario where real-world constraints are taking place, getting larger amounts of synthetic data is just easier and faster. By examining \ref{fig:map_vs_size}(b) we can see that when we train the network with data that was taken directly from the same distribution as the test set distribution, not surprisingly the performance is high. We consider this performance as our upper bound of performance.



\subsection{Training with Mixed Data}
In the case of training with a mixture of manually-gathered and synthetic data, we observed an improvement on both evaluation sets compared to training with synthetic data or manually gathered data alone, as shown in Figures \ref{fig:main_results} and \ref{fig:mix_fig}. Synthetic data might be useful even when manually-gathered data is available. The opposite holds as well: even very small amounts of manually gathered data can help to improve the synthetic data results.

\subsection{Real vs. Synthetic Object Placement }
The real world distribution of the number of items, the occlusions, the size of the objects and their orientations per image is an unknown. The target domain for our dataset is an office setting, but each office in the world is unique. Ideally, we'd want to generate data with office space priors at scale. For this specific task the semantic placement of office space objects at scale was unreasonable. 

To deal with the lack of valid priors, the placement and rotation of the objects were completely randomized. This ensured that no spatial prior was learned by the network due to a bias of the simulated data. Instead of aiming for the common case found in office settings, this data simulates the most challenging visual cases (including scale variation, occlusions, lighting variation, high object density and similar object classes) one could expect in the real-world. Through experiments such as this, we see that generating photorealistic simulated data in extremely visually challenging settings is the key for training robust algorithms that work even in edge cases.

\subsection{Limitations}
This approach may experience difficulties if our examples from our target class, in our case the Expo Markers, change their visual appearance. For example, if they are broken, become dirty, or have their caps removed. For this reason, the main challenge of this method is to anticipate edge cases for our target class. Due to our limited ability to define all edge cases, we recommend always taking a healthy iterative approach, solving the base problem and later moving to additional more complex scenarios and cases.

Another challenge we see is reliably modeling all of the variation in the scene. In this data set, variations in lighting, 3D spatial variations, background variations, number of objects and object classes inserted into the scene were strongly varied. Additional variations to consider for further testing are lighting color, directional emissive object lighting, motion blur and camera noise. 

\say{all models are wrong, but some are useful}
\\[5pt]
\leftline{{\rm--- George E. P. Box}}

This is a key takeaway. It is impossible to perfectly simulate the real world, but it is very possible to create useful simulations to train computer vision neural networks.

\section{Conclusion}
By creating a highly challenging and varied photorealistic synthetic data in 3D environments and successfully training neural networks with it, we gave an empirical evidence that using photorealism in addition to variance, is useful for real world applications. This is the case even when training exclusively on synthetic data.

We believe that our approach can be implemented and generalized to a wide range of objects perception tasks and have an impact in use-cases such as production and assembly lines in smart factories, standard items on shelves of smart stores, food products in smart refrigerators and visual understanding for robots and drones.

\section{Future Work}

\subsection{Additional Testing}

Additional tests we plan to perform in future versions of this paper:

\begin{itemize}
  \item Testing larger datasets is something that we see as important to provide stronger guidelines on the effect of dataset size on trained network quality.

 \item Testing additional types of variations in the data generation pipeline holds promises. Every variation added thus far has improved robustness of the model. 

 \item Testing various neural network backbones on the different size datasets will show whether larger networks are required to utilize the information generated by the synthetic data.
\end{itemize}

 \subsection{Additional Modalities}
Using synthetic data allows us to generate many kinds of labels, and opens a new door to solve tasks that weren't possible for the reason of lacking data. Label types can be divided as follows: 
 \begin{itemize}
  \item 2D - segmentation, sub-segmentation, bounding boxes and key-points.
  \item 3D - depth map, normal map, 3d key-points and specific sensors can be modeled.
  \item 4D - optical flow.
\end{itemize}
 In our next release, we plan to add more types of labels to our images such as key-points, depth map and surface normal map.

\subsection{Domain Adaptation}
  We plan to use noise transfer techniques that attempt to simulate the noise of a target domain on top of our simulated image dataset, in order to minimize the domain gap without affecting the content of the image.

{\small
\bibliography{datagen_main}
}
\clearpage
\appendix
\section{Additional Figures}


\begin{figure*}[ht!]
\centering
\subcaptionbox{}%
  {\includegraphics[width=.39\linewidth]{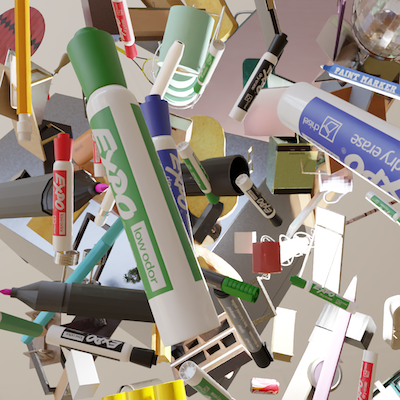}}
\subcaptionbox{}%
  {\includegraphics[width=.39\linewidth]{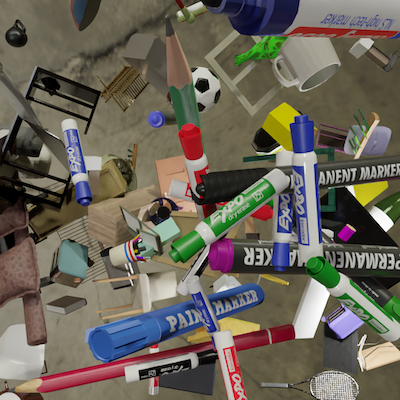}}
\subcaptionbox{}%
  {\includegraphics[width=.39\linewidth]{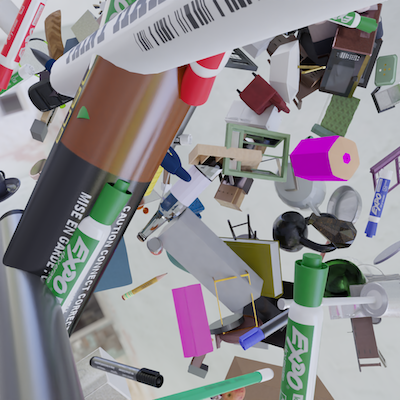}}
\subcaptionbox{}%
  {\includegraphics[width=.39\linewidth]{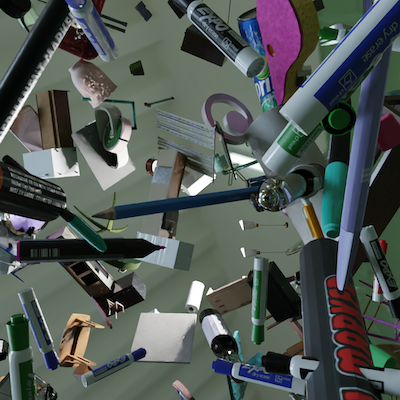}}
\subcaptionbox{}%
  {\includegraphics[width=.39\linewidth]{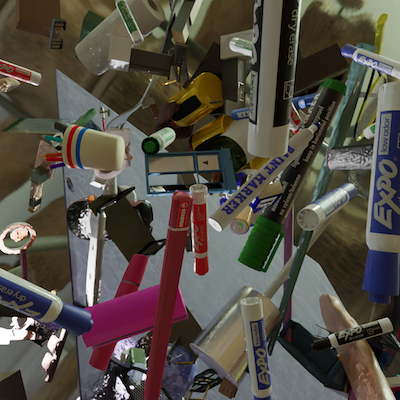}}
\subcaptionbox{}
  {\includegraphics[width=.39\linewidth]{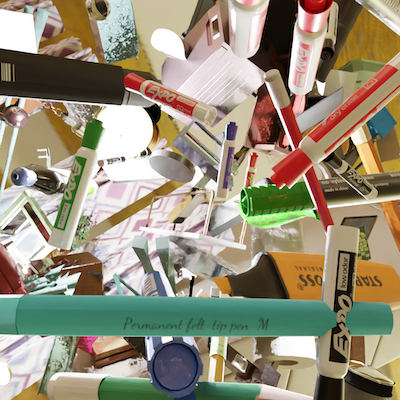}}
 \caption{Samples from $Synthetic$.}
\label{fig:synt_samples}
\end{figure*}

\begin{figure*}[ht!]
\centering
\subcaptionbox{}%
  {\includegraphics[width=.45\linewidth]{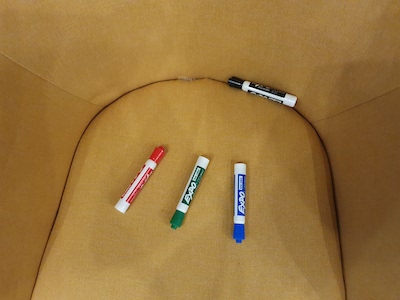}}
\subcaptionbox{}%
  {\includegraphics[width=.45\linewidth]{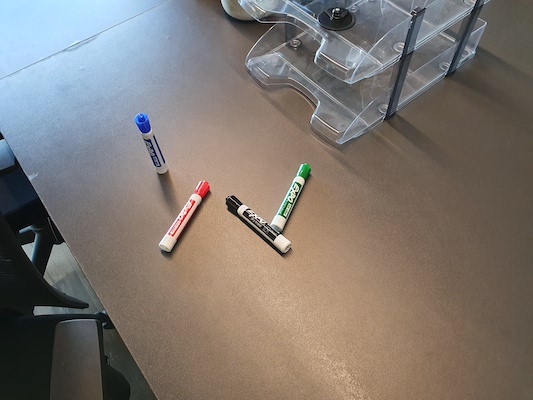}}
\subcaptionbox{}%
  {\includegraphics[width=.45\linewidth]{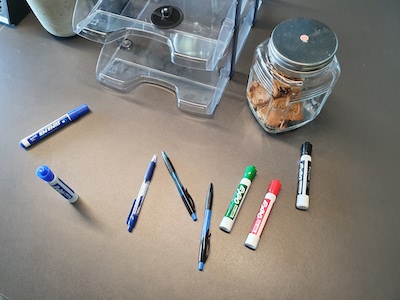}}
\subcaptionbox{}%
  {\includegraphics[width=.45\linewidth]{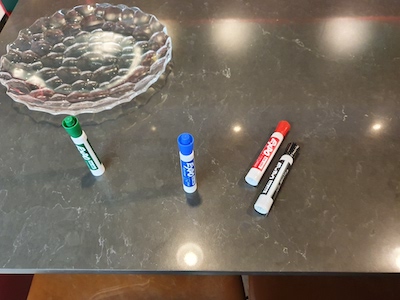}}
\subcaptionbox{}%
  {\includegraphics[width=.45\linewidth]{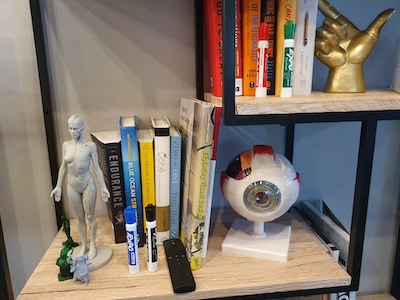}}
\subcaptionbox{}%
  {\includegraphics[width=.45\linewidth]{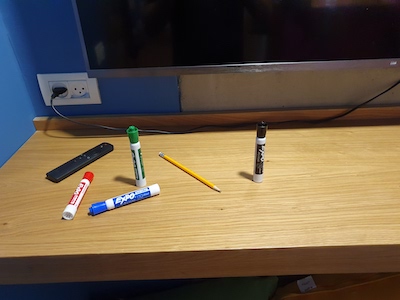}}
 \caption{Samples from $Real\ A$.}
\label{fig:office_samples}
\end{figure*}

\begin{figure*}[ht!]
\centering
\subcaptionbox{}%
  {\includegraphics[width=.45\linewidth]{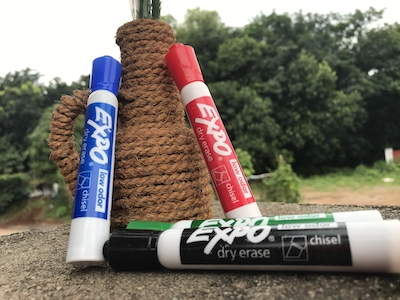}}
\subcaptionbox{}%
  {\includegraphics[width=.45\linewidth]{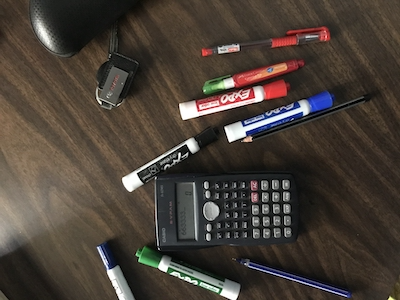}}
\subcaptionbox{}%
  {\includegraphics[width=.45\linewidth]{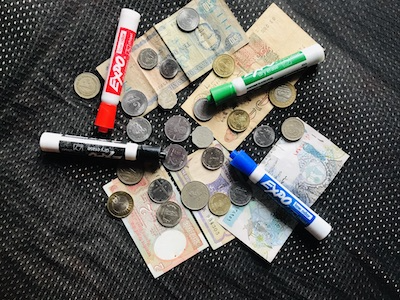}}
\subcaptionbox{}%
  {\includegraphics[width=.45\linewidth]{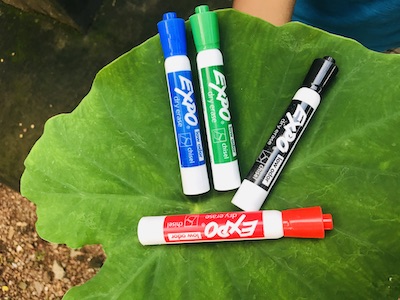}}
\subcaptionbox{}%
  {\includegraphics[width=.45\linewidth]{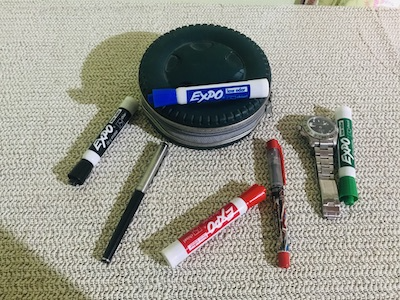}}
\subcaptionbox{}%
  {\includegraphics[width=.45\linewidth]{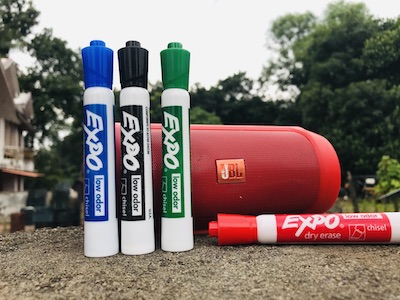}}
 \caption{Samples from $Real\ B$.}
 \label{fig:india_samples}
\end{figure*}

\end{document}